\pdfoutput=1
\documentclass[11pt,onecolumn]{cleantechnicalreport}

\usepackage[authoryear,sort&compress,round]{natbib}
\providecommand{\doi}[1]{doi:\space{\urlstyle{rm}\href{https://doi.org/#1}{\nolinkurl{#1}}}}
\usepackage{hybridfrontmatter}
\usepackage{nicefrac}
\usepackage{flafter}
\usepackage{placeins}
\usepackage{float}
\usepackage{makecell,multirow}
\usepackage{longtable}
\usepackage[noabbrev]{cleveref}

\usepackage{fancyhdr}
\usepackage{xcolor}
\usepackage{tcolorbox}

\definecolor{mybg_yellow}{RGB}{245, 241, 233}
\definecolor{mybg_blue}{RGB}{242, 243, 245}

\newtcolorbox{targetbox_yellow}{
    colback=mybg_yellow,       
    colframe=mybg_yellow,      
    arc=3pt,            
    boxrule=0pt,        
    left=10pt,          
    right=10pt,         
    top=8pt,            
    bottom=8pt,         
    breakable,          
    enhanced,           
    before skip=10pt,   
    after skip=10pt     
}
\newtcolorbox{targetbox_blue}{
    colback=mybg_blue,       
    colframe=mybg_blue,      
    arc=3pt,            
    boxrule=0pt,        
    left=10pt,          
    right=10pt,         
    top=8pt,            
    bottom=8pt,         
    breakable,          
    enhanced,           
    before skip=10pt,   
    after skip=10pt     
}

\title{Turing Agent: Agentic Formal Verification for\\ Mathematical Research}

\renewcommand{\ReportAuthorFont}{\normalfont\fontsize{10.5}{14}\selectfont}
\author{Yang Chen\textsuperscript{\normalfont$1,\star$}}
\author{Qianyi Wang\textsuperscript{\normalfont$1,\star$}}
\author{Ronghui Yang\textsuperscript{\normalfont$1,\ast$}}
\author{Zhanhao Li\textsuperscript{\normalfont$1,\ast$}}
\author{Yu Cong\textsuperscript{\normalfont$1,2$}}
\author{Chen Zhang\textsuperscript{\normalfont$1$}}
\author{Hangfan Zhang\textsuperscript{\normalfont$1$}}
\author{Shao Zhang\textsuperscript{\normalfont$1$}}
\author{Shuai Li\textsuperscript{\normalfont$3$}}
\author{Toru Takisaka\textsuperscript{\normalfont$2$}}
\author{Chao Xu\textsuperscript{\normalfont$2$}}
\author{Yi Zhou\textsuperscript{\normalfont$2$}}
\author{Zhihao Tang\textsuperscript{\normalfont$4$}}
\author{Xiao Wang\textsuperscript{\normalfont$4$}}
\author{Bo Zhang\textsuperscript{\normalfont$1,\dagger$}}
\author{Shuyue Hu\textsuperscript{\normalfont$1,\dagger$}}

\affil{\textsuperscript{\normalfont$1$}Agent Center, Shanghai Artificial Intelligence Laboratory, \\
\textsuperscript{\normalfont$2$}Algorithms and Logic Group, University of Electronic Science and Technology of China, \\
\textsuperscript{\normalfont$3$}School of Computing and Data Science, The University of Hong Kong, \\
\textsuperscript{\normalfont$4$}Institute for Theoretical Computer Science, Shanghai University of Finance and Economics. \\
\textsuperscript{$\star$}Lead Author, \textsuperscript{$\ast$}Core Contributor, 
\textsuperscript{$\dagger$}Project Lead \\
{\bf Contact:} \url{hushuyue@pjlab.org.cn}, \url{chenyang4@pjlab.org.cn}, \url{wangqianyi@pjlab.org.cn}\\
{\bf Website:} \url{https://wangqy031216-cloud.github.io/turing-agent}
}
\DeclareUnicodeCharacter{2212}{-}

\begin{abstract}
Large language models (LLMs) are rapidly advancing mathematical discovery, moving beyond benchmark-level problem solving to tackle frontier research problems. This progress is ushering mathematics into an era of proof abundance, raising a fundamental challenge: how can we ensure the correctness of the ever-growing volume of AI-generated proofs and transform them into trustworthy, machine-verified mathematical knowledge?
We introduce \emph{Turing Agent}, an agent harness designed to orchestrate end-to-end formal verification of research-level mathematics. 
 Its \emph{two-scale} architecture integrates theorem formalization and proof formalization agents, empowered by structured proof decomposition, parallel verification, accumulated proof experience, and selective semantic review. 
 Leveraging our collaborations with theoretical computer science (TCS) researchers, we focus our evaluation on TCS, a particularly demanding domain characterized by intricate quantification, complex dependencies, and stringent semantic requirements.
 Against direct generation and general-purpose coding harnesses, including Codex CLI and Claude Code, Turing Agent achieves approximately \textbf{57--62\% faithful formalization success} on Formal TCS Bench and \textbf{83\% overall pass@1} on a TCS-focused subset of RLMEval, using GLM-5.3 and DeepSeek-V4-Pro. 
 Notably, Turing Agent, built on DeepSeek, successfully completes the formal verification of three \textbf{previously unresolved } TCS tasks from OpenAI's October 2026 collection.
 One particularly challenging TCS problem required about a hundred formally verified auxiliary claims. 
Moreover, by actively exploring and refining the formalization process, Turing Agent can uncover simpler alternative proofs, expose unnecessary dependencies in original proofs, and generate reusable mathematical knowledge as a byproduct of verification. 
\end{abstract}

\begin{document}
\vspace*{-1.9cm}
\noindent\hspace*{-3pt}\includegraphics[height=0.8cm]{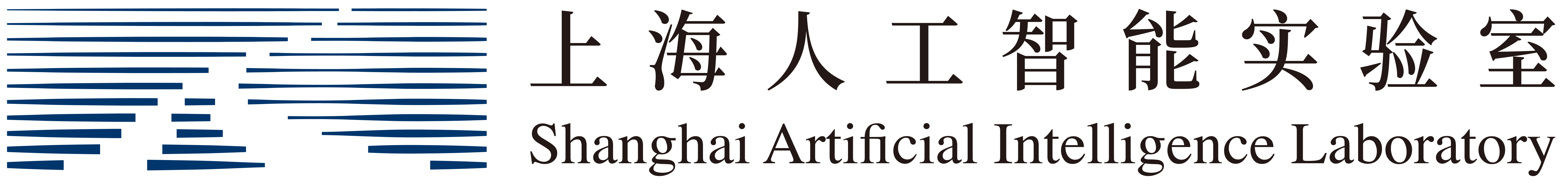}\par\vspace{1.2em}
\maketitle
\suppressfloats[t]

\begin{figure}[t]
    \centering
    \includegraphics[width=\linewidth]{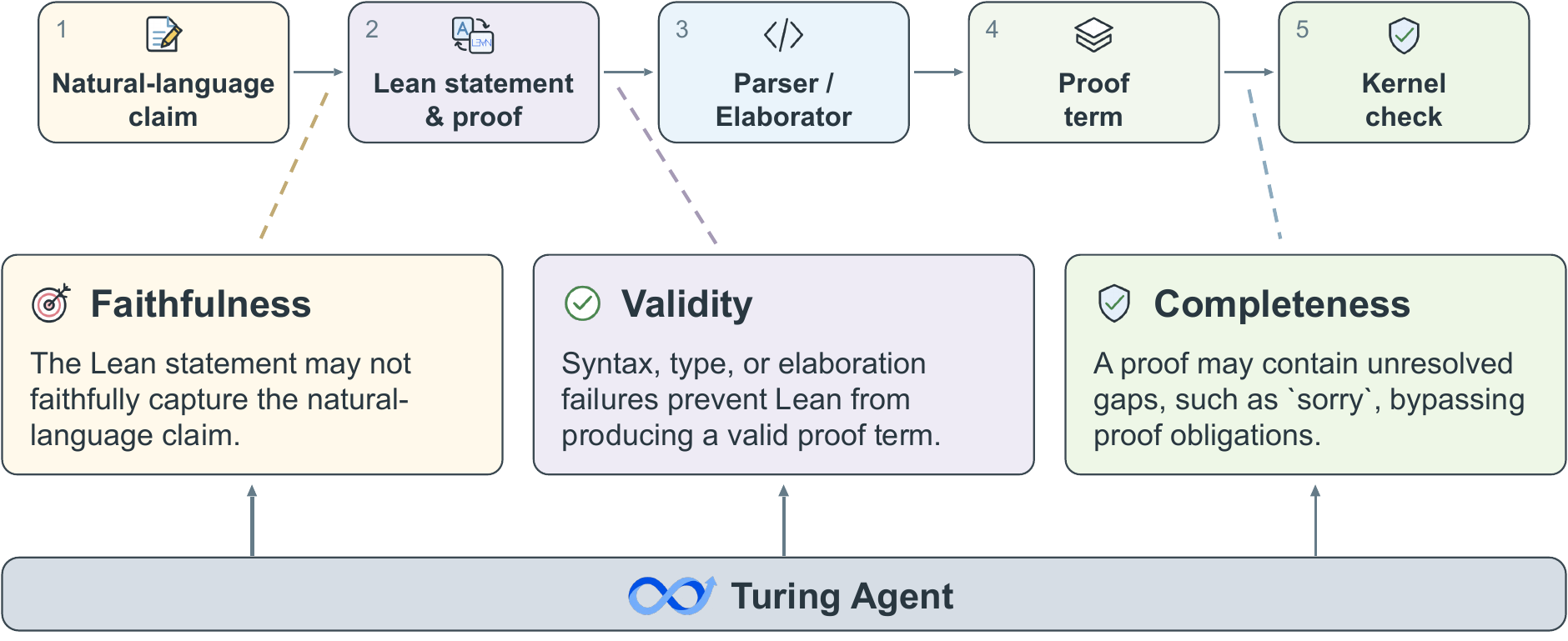}
    \caption{Three challenges in automated formal verification. The formal statement must preserve the original claim (faithfulness), the code must pass Lean’s syntax and type checks (validity), and the proof must discharge all obligations without gaps (completeness). Our Turing Agent is designed to satisfy these requirements throughout the formalization process.}
    \label{fig:formalization-challenges}
\end{figure}

\section{Introduction}
Large language models (LLMs) are fundamentally reshaping the process of mathematical discovery. Moving beyond competition problems, LLM-based AI systems are increasingly making breakthroughs on longstanding research challenges, including solving previously open Erdős problems~\citep{feng2026erdos}, disproving the Erdős unit-distance conjecture~\citep{openai2026unitdistance}, and tackling the celebrated Navier--Stokes existence and smoothness problem~\citep{openai2026navierstokes}. As Terence Tao envisions in his conception of Math 2.0, we are approaching a fundamental transition from an era of proof scarcity to proof abundance \citep{tao2026world}. As AI systems generate mathematical proofs at unprecedented scale and speed, a fundamental question arises: who will verify the correctness of this ever-growing volume of AI-generated proofs?  In this emerging era, verification is becoming one of the defining bottlenecks of mathematical discovery---not merely for checking individual proofs, but for transforming the abundance of AI-generated results into trustworthy mathematical knowledge.

Formal verification offers a principled and scalable foundation for addressing this bottleneck: proof assistants such as Lean can turn mathematical claims and proofs into machine-checkable objects certified by a small trusted kernel~\citep{demoura2015lean}. Yet verifying realistic research artifacts involves substantially more than proving an isolated theorem whose formal statement has already been provided. Research papers introduce their own definitions, conventions, assumptions, intermediate results, and dependencies, all of which must be translated faithfully before proof search can even begin. A Lean statement may compile while subtly changing the domain, quantifiers, assumptions, or intended semantics of the original claim, so syntactic validity alone does not guarantee semantic faithfulness~\citep{zhang2026beyond,han2026shadowbench,meek2026formalizing}. Similar risks arise during proof construction, where auxiliary lemmas, representation choices, and local simplifications can drift from the intended argument or create long-range dependencies across a large proof development.

\begin{figure}[t]
    \centering
    \includegraphics[width=\linewidth]{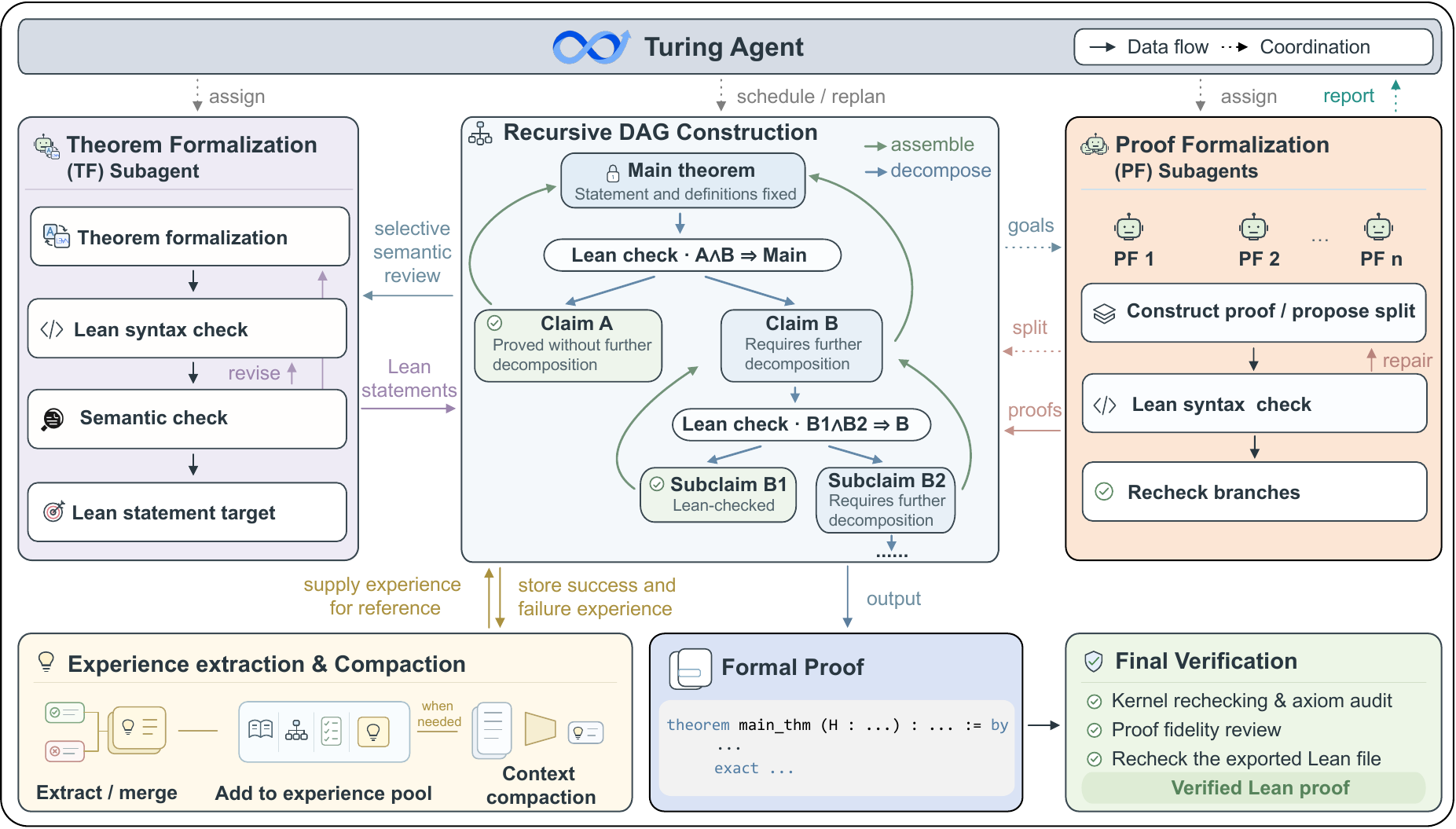}
    \caption{Turing Agent coordinates theorem formalization (TF) and proof formalization (PF) agents. TF agents provide checked Lean statements and definitions, while PF agents construct proofs and request semantic review when needed. The resulting proof undergoes final verification. In its runtime, Turing Agent coordinates recursive DAG construction, parallel proof formalization, and selective semantic review. Experience from successful and failed attempts is extracted and reused, while checked branch proofs are assembled for final verification.}
    \label{fig:workflow}
\end{figure}

These challenges, illustrated in Figure~\ref{fig:formalization-challenges}, motivate a broader view of automated formal verification as an \emph{agent harness} problem rather than only a theorem-proving problem. Most recent automated theorem proving systems, such as \emph{LeanDojo/ReProver}~\citep{yang2023leandojo}, \emph{DeepSeek-Prover-V1.5}~\citep{xin2024deepseekprover}, \emph{AlphaProof}~\citep{hubert2026alphaproof}, and \emph{Goedel-Prover-V2}~\citep{lin2025goedelv2}, primarily focus on proof search from already formalized theorem statements. Research-level verification requires coupling theorem formalization with semantic checking to preserve \textbf{\emph{faithfulness}} to the original claim. Iterative debugging must resolve syntax, type, and elaboration errors to ensure \textbf{\emph{validity}} of the Lean code. Across long proof developments, proof decomposition, dependency management, and reuse of intermediate results must also be coordinated to achieve \textbf{\emph{completeness}}, with all proof obligations discharged and no unresolved gaps.


Motivated by these challenges, we introduce \textbf{Turing Agent}, an agent harness designed to transform informal mathematical research into faithful, machine-checked formal developments. Rather than treating formal verification as isolated proof search, Turing Agent orchestrates the entire verification process, from understanding research-level mathematical statements to constructing, validating, and integrating complex proof developments. As illustrated in Figure~\ref{fig:workflow}, its \emph{two-scale} architecture coordinates a {\em theorem formalization (TF) agent} and a {\em proof formalization (PF) agent}, grounded in formal libraries such as Mathlib~\citep{mathlib2020} and domain-specific resources such as CSLib~\citep{barrett2026cslib}. The TF agent translates informal mathematical claims into semantically faithful Lean statements, while the PF agent builds complete proofs through structured decomposition and iterative feedback from the proof assistant. Crucially, Turing Agent treats large-scale verification as a coordinated, evolving process: a directed acyclic graph (DAG) organizes proof obligations and unlocks parallel exploration, an experience pool accumulates reusable lemmas and lessons from failed attempts, and selective semantic review safeguards critical intermediate claims against semantic drift. By integrating these capabilities into a single verification workflow, Turing Agent goes beyond generating proofs for individual statements toward systematically transforming complex research arguments into trustworthy, reusable, machine-verified artifacts.

\begin{figure}[t]
    \centering
    \includegraphics[width=\linewidth]{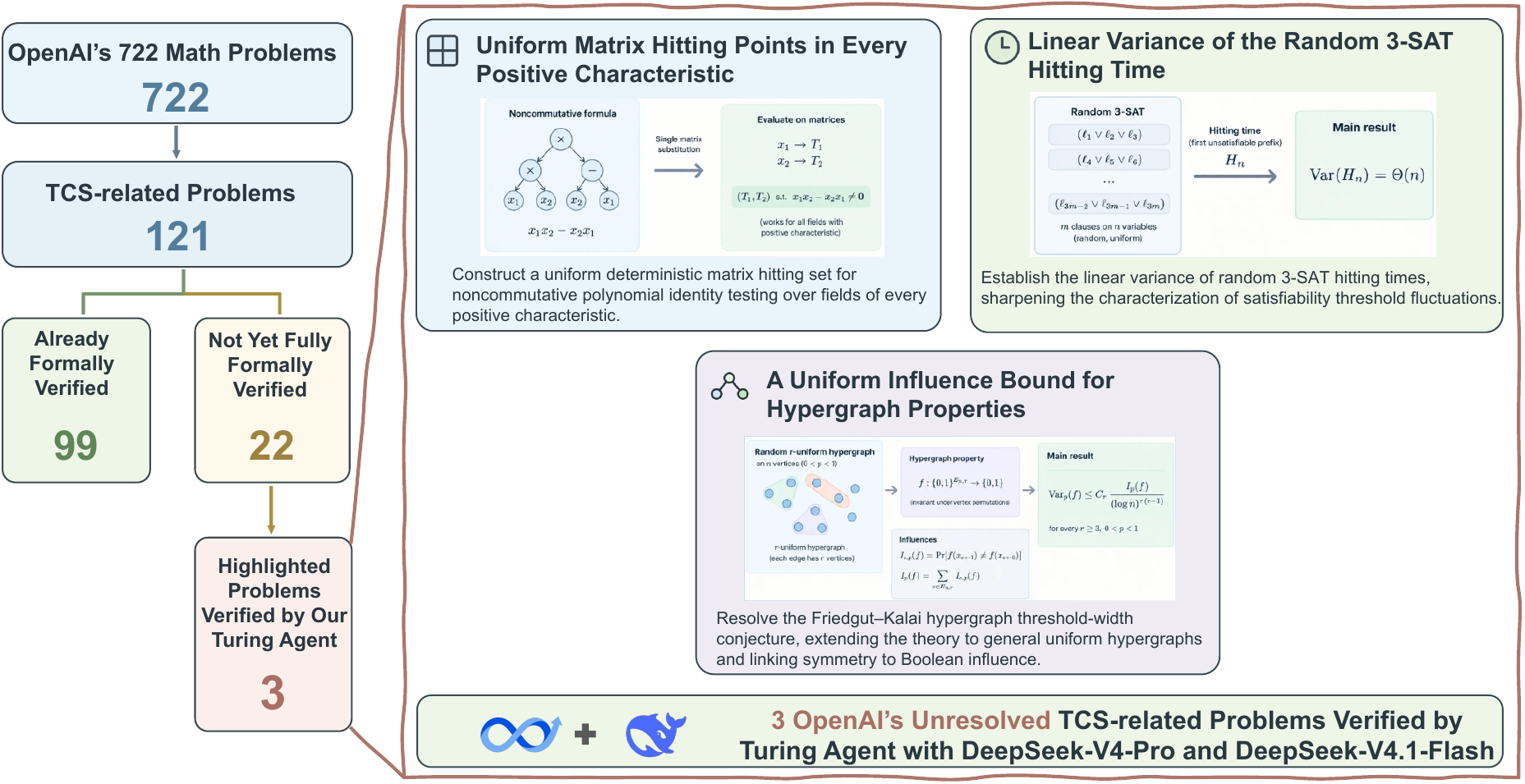}
    \caption{Research-level TCS verification beyond standard benchmarks. Of the 121 TCS problems in OpenAI’s mathematical research collection, 22 had not been fully verified in Lean. Among these, we highlight three examples formally verified by Turing Agent using DeepSeek-V4-Pro and DeepSeek-V4.1-Flash.}
    \label{fig:openai-verification-overview}
\end{figure}

We demonstrate the effectiveness of Turing Agent across two complementary dimensions of formal verification: \emph{faithful theorem formalization} and \emph{automated proof construction}. 
We focus our evaluation on theoretical computer science (TCS), which presents distinctive challenges involving intricate quantification, complex parameter dependencies, and reasoning over large or infinite classes of instances. These challenges, together with our collaboration with TCS researchers who provide domain expertise for assessing semantic faithfulness, make TCS a particularly suitable testbed for evaluating research-level formal verification. Accordingly, we select \emph{Formal TCS Bench}~\citep{wang2026formaltcs} for theorem formalization and a TCS-focused subset of \emph{RLMEval}~\citep{poiroux2025rlmeval} for proof construction. We compare Turing Agent against direct model generation and two widely used general-purpose coding harnesses, \textbf{Codex CLI} and \textbf{Claude Code}, using the same underlying models, \textbf{GLM-5.3} and \textbf{DeepSeek-V4-Pro}. On the 143 problems in Formal TCS Bench, Turing Agent achieves near-complete compilation and approximately \textbf{57--62\% semantic success}, substantially outperforming the strongest baseline in faithful theorem formalization. On the 248-theorem TCS subset of RLMEval, Turing Agent achieves approximately \textbf{83\% overall pass@1}, demonstrating strong automated proof-construction capability. The substantial gap between compilation and semantic success further underscores the importance of preserving mathematical meaning rather than merely producing syntactically valid Lean statements. Together, these results demonstrate how harness-level coordination can significantly extend the verification capabilities of underlying language models beyond those of general-purpose coding agents.

More compellingly, we take Turing Agent beyond established benchmarks to tackle \emph{research-level formal verification tasks that frontier AI systems had previously left incomplete}. We investigate problems from OpenAI's collection of AI-generated mathematical research manuscripts~\citep{openai2026math}, focusing on theoretical computer science out of the same motivation as our benchmark evaluation. 
Instead of repeating formal verifications already accomplished by the original system, we deploy Turing Agent powered by DeepSeek-V4-Pro and DeepSeek-V4.1-Flash to complete verification tasks that remained unresolved in the original developments. Figure~\ref{fig:openai-verification-overview} summarizes the task selection and highlights three problems addressed by Turing Agent. We highlight {\bf three} representative cases spanning distinct research areas: \textit{uniform matrix hitting for noncommutative formulas in algebraic complexity}, \textit{a uniform influence bound for hypergraph properties in probabilistic combinatorics}, and \textit{a linear variance bound for the random 3-SAT hitting time}. Completing these verifications requires substantially more than local proof repair: Turing Agent must reconstruct missing mathematical arguments, introduce and verify new auxiliary lemmas, navigate specialized formal libraries, and coordinate extensive dependencies across long proof developments. In the more challenging cases, this process produces dozens to over one hundred verified auxiliary claims, culminating in complete root theorems certified by Lean's trusted kernel. These results demonstrate that a carefully designed agent harness can extend the verification frontier of an underlying language model, completing research-level formal proofs left unresolved by previous AI-driven verification efforts.

Beyond completing previously unresolved proofs, our experiments reveal two interesting behaviors of agents during formal verification. {\em First, agents may choose different mathematical constructions from those used in the original proofs to simplify subsequent reasoning.} When proving an existential statement, multiple constructions may satisfy the required conditions. Rather than following the original proof exactly, Turing Agent may select a different construction that makes the remaining proof obligations easier to discharge. Take \(C=\max(A,B)\), for an illustrative example: instead of choosing it as in the original proof, the agent chooses \(C=A+B+1\) for \(A,B\geq 0\), allowing subsequent inequalities to be handled directly by linear arithmetic. 
{\em Second, agents may generate and verify auxiliary lemmas that are ultimately unnecessary for the final proof.} We observe that Turing Agent constructs and verifies intermediate nodes in its proof dependency graph, but some of these nodes are not used in the final proof of the main theorem. These unused nodes remain formally verified, even though they are not required to establish the final result. Based on these observations obtained from Turing Agent, we suggest two promising directions for designing agentic formal verification: allowing agents to adapt mathematical constructions rather than rigidly reproduce original proofs, and improving proof planning to identify and prioritize dependencies that are essential to the final theorem. 

The remainder of this report is organized as follows. Section~\ref{sec:method} introduces the design of Turing Agent and its formal verification workflow. Section~\ref{sec:eval} presents evaluations on theorem formalization and proof formalization benchmarks, together with comparisons against general-purpose coding harnesses and direct model generation. Section~\ref{sec:open} studies research-level verification cases beyond standard benchmarks. Section~\ref{sec:conclusion} discusses limitations, future directions, and concludes the report.

\section{Related Work}
\label{sec:related-work}

\subsection{Agents for Problem Solving in Natural Language}

Mathematical agents increasingly organize problem solving around repeated proof generation, critique, and revision. {\em Aletheia} combines this process with literature search for research-level problems~\citep{feng2026aletheia}. {\em Rethlas} augments a generation-and-review loop with theorem retrieval, examples, counterexamples, and recursive decomposition~\citep{ju2026conjecture}.

Longer arguments require tracking established claims and unresolved difficulties across many proof attempts. {\em Danus} coordinates parallel workers through a shared fact graph whose entries are reviewed by an independent natural-language verifier~\citep{liu2026danus}. {\em RMA} stores proof revisions, unresolved issues, and literature records to prepare context for individual reasoning tasks~\citep{zhao2026rma}. {\em Ansatz} extends graph memory to facts, plans, and counterexamples, with retrieval and experience reuse across problems~\citep{zhang2026continual}. Danus and Ansatz accept intermediate claims through language-model review, which is distinct from formal proof checking.

\subsection{Agents for Formal Verification}

Formal reasoning systems use proof-assistant feedback to guide proof construction. {\em LeanDojo/ReProver} augments proof search with premise retrieval, while {\em AlphaProof} learns proof-search policies through reinforcement learning~\citep{yang2023leandojo,hubert2026alphaproof}. Agent systems combine formal checking with informal reasoning and external tools. {\em Aristotle} combines formal search with informal lemma generation~\citep{achim2025aristotle}, while {\em Hilbert} recursively decomposes difficult claims~\citep{varambally2025hilbert}. {\em Numina}-Lean-Agent equips a general coding agent with Lean interaction, theorem retrieval, and auxiliary reasoning tools~\citep{liu2026numina}.

Research-level formalization also requires defining concepts and checking their correspondence to the source text. {\em Archon} and {\em Theo} coordinate planning, theorem retrieval, definition development, and iterative Lean formalization~\citep{ju2026conjecture,moakhar2026theo}. Theo additionally tests proposed definitions through auxiliary lemmas. {\em MMAT} combines natural-language and formal provers with knowledge management and explicit task dependencies~\citep{cao2026mechmath}. {\em AIProver} jointly improves its model and agent workflow using feedback on formal and semantic correctness~\citep{jana2026aiprover}.

\subsection{Agent Evaluation}

Natural-language proofs are evaluated through expert assessment and model-based review of the argument. {\em Aletheia} uses expert judgments of whether research proofs are publishable after minor revisions~\citep{feng2026aletheia}. {\em TCS-Bench} uses an LLM verifier that receives the target statement, context, and reference proof, and calibrates its judgments against expert-labeled examples~\citep{cohenaddad2026tcsbench}. Step-level verification examines individual deductions, tracks local assumptions and cited results, and separates unresolved steps from demonstrated errors~\citep{sun2026strict}.

Formal proof evaluation checks whether Lean accepts a complete proof of the supplied statement, as in {\em RLMEval}~\citep{poiroux2025rlmeval}. This establishes correctness for that statement, while evaluating statement formalization requires checking its correspondence to the original claim. {\em BEq+} compares a candidate with a reference statement by attempting to prove each from the other through several Lean tactic strategies~\citep{poiroux2025reliable}. Failure to find these proofs can reflect limited search rather than an incorrect formalization.

Other methods assess semantic alignment through auxiliary statements or direct review.  ShadowBench's SA-PASS uses auxiliary statements, called {\em shadows}, that together characterize a reference claim. Evaluation requires the candidate to imply every shadow and all shadows together to imply the candidate~\citep{han2026shadowbench}. When no reference Lean statement is available, Beyond Compilation combines compilation with agreement between two LLM judges that compare the formalization with the source text, and validates this criterion through human review~\citep{zhang2026beyond}.

\section{Design of the Framework}
\label{sec:method}

Faithful statement formalization is essential to verifying the intended theorem in Lean. Yet evaluations of existing agents have reported errors that change the proof target~\citep{zhang2026beyond,han2026shadowbench,meek2026formalizing}, and a proof of such a target may pass Lean verification without establishing the original theorem. Turing Agent therefore adopts a \emph{two-scale design} with distinct agents for theorem formalization and proof formalization, as illustrated in Figure~\ref{fig:workflow}. Theorem formalization provides Lean statements and supporting definitions that pass both Lean checking and semantic review. Proof formalization uses these statements as proof targets and constructs Lean proofs from the supplied informal arguments. The main theorem must be formalized and reviewed by a theorem formalization agent before proof construction begins. Intermediate claims undergo the same procedure selectively during proof construction.

\subsection{Theorem Formalization}
\label{sec:statement-translation}

Theorem formalization can run independently or be invoked during proof construction. Given an informal theorem statement and its mathematical context, the {\bf theorem formalization agent (TF agent)} searches Mathlib~\citep{mathlib2020} and CSLib~\citep{barrett2026cslib} for suitable formal definitions. The agent inspects the retrieved definitions and introduces new ones for concepts not covered by the libraries, then uses these definitions to construct the Lean statement.

The TF agent first checks the statement and its supporting definitions with Lean. Once this check passes, the agent reviews whether the formalization preserves the original domains, assumptions, quantifier dependencies, and conclusion. This semantic review compares the original text with the Lean statement, its elaborated type, and relevant definitions, including the full bodies of newly introduced definitions. If either check reveals a problem, the TF agent revises the statement or definitions and repeats the checks. Once both checks pass, the agent returns the formal statement and its supporting definitions. For the main theorem, these remain fixed throughout proof construction.

\subsection{Proof Formalization}
\label{sec:proof-translation}
Given a Lean statement and its corresponding informal proof, {\bf proof formalization agents (PF agents)} construct a Lean proof by following the informal argument and decomposing the proof into intermediate claims when needed. Turing Agent organizes these claims and their dependencies in a {\bf directed acyclic graph (DAG)} and coordinates their proof construction, as illustrated in Figure~\ref{fig:workflow}.

\subsubsection{Proof Decomposition and Parallel Construction}

Guided by the informal proof, a PF agent decomposes a parent goal into intermediate claims when needed and records their dependencies in the DAG. Before pursuing these claims, the PF agent must show in Lean that their formal statements, taken as assumptions, imply the complete parent proposition. Turing Agent accepts the decomposition only after Lean verifies that the child statements imply the parent statement. The same procedure is applied recursively to the child claims. If a goal can be proved directly, the PF agent submits the proof without splitting the goal further.

Once the decomposition is activated, Turing Agent can assign independent child goals to multiple PF agents concurrently. Each PF agent uses results from Mathlib and CSLib to construct the assigned proof. Based on Lean feedback, the agent revises the proof or further decomposes the claim as needed. Turing Agent rechecks completed branch proofs before using them to construct the parent proof. Once the child claims are proved, it uses the checked derivation to complete the parent proof. This process continues upward until the main theorem is reached.

\subsubsection{Selective Semantic Review}

The decomposition check establishes that the child claims imply their parent, but does not check their correspondence to the informal argument. Misinterpreted definitions or assumptions can therefore misdirect subsequent proof construction. Checking semantic alignment for every intermediate claim would require repeated model calls, increasing both runtime and token consumption. We therefore review intermediate claims selectively to address semantic errors while limiting these costs.

\begin{figure}[!h]
    \centering
    \includegraphics[width=0.7\linewidth]{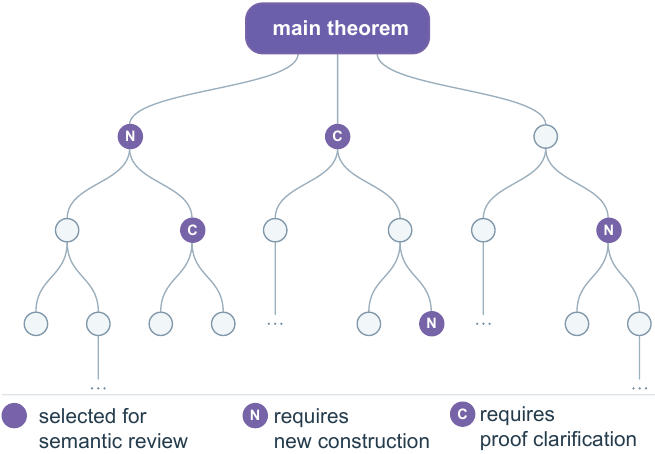}
    \caption{Selective semantic review in the proof DAG. PF agents request TF agents when a concept is unclear or a new definition is introduced. The TF agent checks whether the statement and its definitions preserve the intended meaning.}
    \label{fig:selective-semantic-review}
\end{figure}

PF agents normally formalize intermediate claims directly, using Lean to check routine deductions and direct applications of library lemmas. PF agents request semantic review when the formal meaning of a new concept is uncertain, or when a claim's assumptions or quantifier dependencies are unclear. They also request review when an added or reformulated claim may change the scope or conclusion of the informal argument. Figure~\ref{fig:selective-semantic-review} illustrates the selection of intermediate claims for review within the proof DAG. For each selected claim, Turing Agent invokes the TF agent to check the statement and its supporting definitions using the procedure in Section~\ref{sec:statement-translation}. The PF agent uses the returned feedback to revise the statement or definitions when needed.

\bigskip

\subsubsection{Experience Extraction and Reuse}

\begin{figure}[!htbp]
    \centering
    \includegraphics[width=0.8\linewidth]{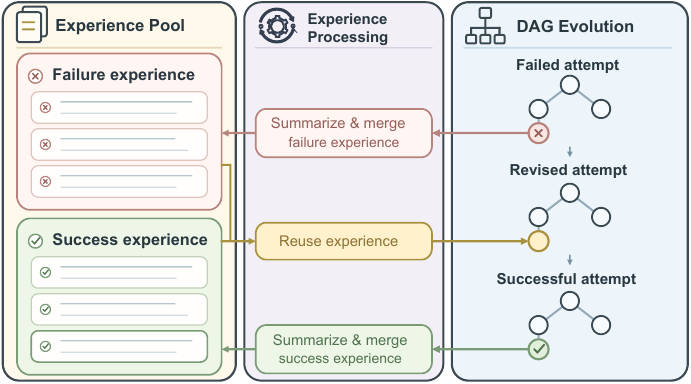}
    \caption{Experience extraction and reuse during DAG evolution. PF agents use retained records to guide revisions and update the pool when later attempts yield useful findings.}
    \label{fig:experience-reuse}
\end{figure}

During proof formalization, compressing the interaction history can discard useful findings from earlier attempts. To preserve them, PF agents selectively extract experience records from successful and failed attempts, using completed checks as evidence. These records capture useful applications of library lemmas, required side conditions, and causes or corrections of failed attempts. When new findings extend an existing record for the same claim, the PF agent merges them into a revised record while retaining the supporting evidence. Figure~\ref{fig:experience-reuse} illustrates this cycle: existing records guide revised proof attempts, and useful findings from successful revisions are retained for later reuse.

The resulting records form an experience pool that keeps useful findings available when the interaction history is compressed. When context compaction is needed, the PF agent reviews recent attempts and saves any useful findings still missing from the pool. Turing Agent then compresses the history and supplies PF agents with records relevant to the current claim and its dependencies, as illustrated in Figure~\ref{fig:workflow}. These records provide a concise context for the next proof attempt.

\subsection{Final Verification}

Once all auxiliary claims have been proved, Turing Agent checks the complete proof with an independent Lean kernel and confirms that it establishes the fixed theorem statement. It then assigns a review agent to compare the main proof and its auxiliary lemmas with the informal argument and supporting source context.

The review agent distinguishes changes in presentation from changes to the mathematical argument. It allows regrouping of steps, explicit treatment of side conditions supported by the source context, and direct use of cited lemmas. It flags omitted essential steps, unsupported assumptions, and substitution of a different argument. Turing Agent returns the Lean proof only after it passes both formal verification and semantic review.

\begin{figure}[t]
    \centering
    \includegraphics[width=\linewidth]{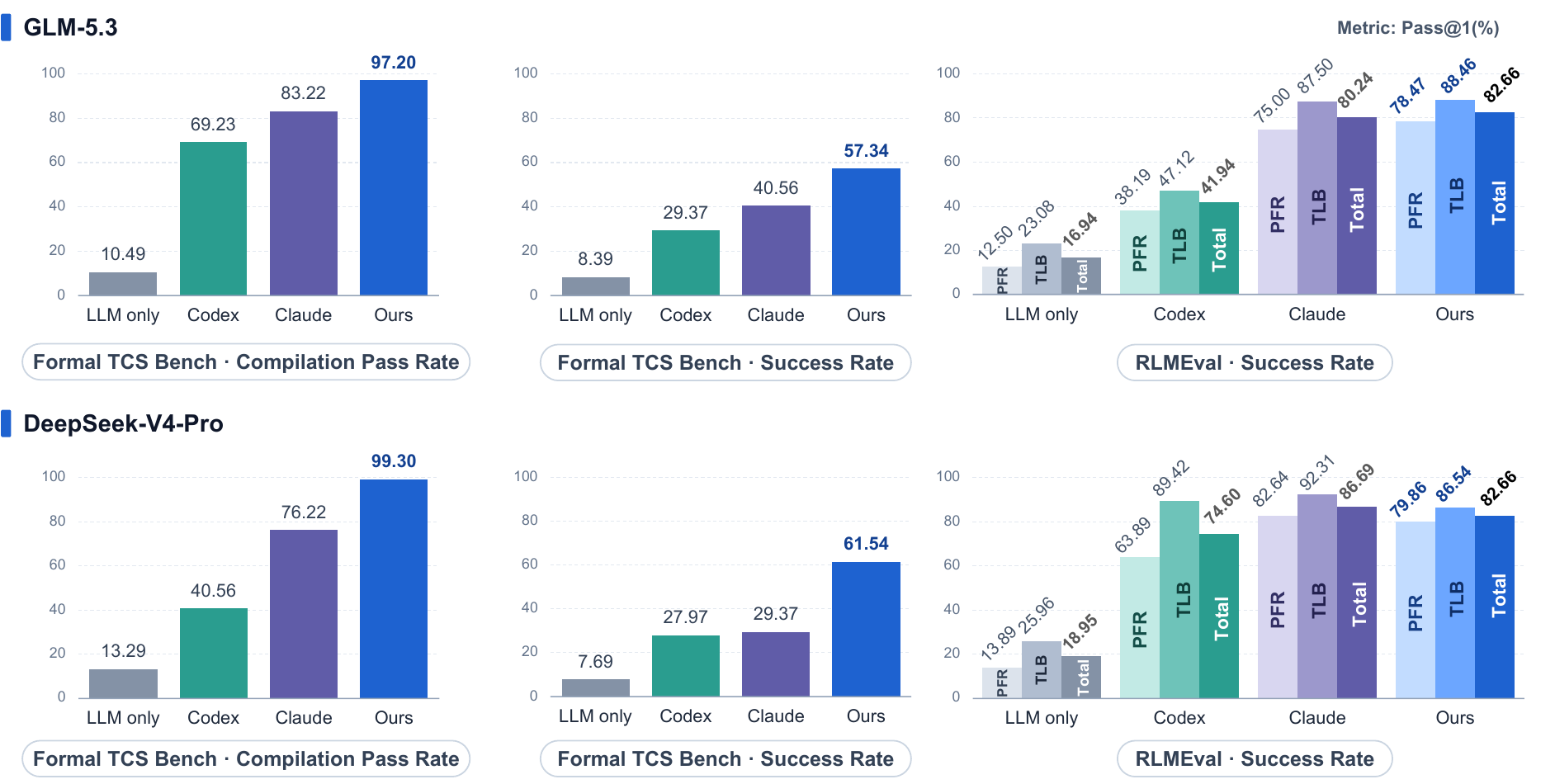}
    \caption{Pass@1 results (\%) on Formal TCS Bench and RLMEval. Formal TCS Bench also reports compilation rates. RLMEval results are shown separately for PFR and TLB and combined as Total. PFR covers additive combinatorics and entropy methods from the Polynomial Freiman–Ruzsa project, with applications to linearity testing in TCS. TLB (Testing Lower Bounds) covers information-theoretic lower bounds for statistical hypothesis testing.}
    \label{fig:benchmark-results}
\end{figure}

\section{Experimental Evaluation}
\label{sec:eval}
\subsection{Setup}
\label{sec:experimental-setup}

\paragraph{Benchmarks.}
We evaluate theorem formalization and proof formalization separately. For theorem formalization, we use the 143 problems in {\bf Formal TCS Bench}~\citep{wang2026formaltcs}, which pairs theorem statements from recent theoretical computer science papers with reference Lean formalizations. The task is to translate each informal statement into a Lean declaration. For proof formalization, we use a 248-theorem evaluation set drawn from the Polynomial Freiman–Ruzsa (PFR) and TestingLowerBounds (TLB) projects in {\bf RLMEval}~\citep{poiroux2025rlmeval}. Each task provides a formal theorem statement, its informal proof, and the corresponding project context. We use the benchmark's easy mode, which makes project-specific auxiliary lemmas available during proof construction.

\paragraph{Models and Baselines.}
We use GLM-5.3~\citep{zai2026glm53} and DeepSeek-V4-Pro~\citep{deepseekai2026v4pro}. For each model, we compare Turing Agent with direct generation, Codex CLI~\citep{openai2025codexcli}, and Claude Code~\citep{anthropic2026claudecode}. These comparisons assess performance across different harnesses while keeping the model fixed.

\paragraph{Evaluation Metrics.}
We report pass@1 using one final submission per problem. An agent may revise its work using Lean feedback within a run; intermediate attempts are not counted as separate submissions. On Formal TCS Bench, we report both the compilation rate and the success rate. The compilation rate measures whether Lean accepts the generated declarations. The benchmark's original metric, BEq+~\citep{poiroux2025reliable}, uses automated proof search to establish equivalence between a candidate statement and its reference. For these difficult theorems, this search may not establish equivalence within the evaluation time limit, even when the translation is correct. We therefore use independent semantic review to assess whether the generated statement and its supporting definitions preserve the meaning of the original theorem. A submission counts as successful only if it compiles and passes this review. Semantic review on Formal TCS Bench is performed using GPT-6.1 Sol~\citep{openai2026gpt61sol}.

On RLMEval, success requires a complete Lean proof of the supplied statement, with no remaining \texttt{sorry}. We report results separately for PFR and TLB, as well as the overall success rate across the evaluation set.

\subsection{Results}
\label{sec:formal-tcs-bench}

\paragraph{Theorem Formalization.}
Turing Agent outperforms all evaluated baselines on Formal TCS Bench with both models (Figure~\ref{fig:benchmark-results}). With GLM-5.3 and DeepSeek-V4-Pro, the success rates reach 57.34\% and 61.54\%, respectively, exceeding Claude Code, the strongest baseline, by 16.78 and 32.17 percentage points. Turing Agent also achieves the highest compilation rates at 97.20\% and 99.30\%, respectively. The gap between compilation and success rates reinforces the need to assess semantic faithfulness separately from Lean acceptance.

\paragraph{Proof Formalization.}
On RLMEval, Turing Agent achieves overall pass@1 rates above 82\% with both models. With GLM-5.3, it obtains the highest score among the evaluated methods, exceeding Claude Code by 2.42 percentage points. With DeepSeek-V4-Pro, it reaches 82.66\%, improving on Codex CLI by 8.06 percentage points and remaining within 4.03 percentage points of Claude Code's 86.69\%. These results complement the gains on Formal TCS Bench, showing that Turing Agent supports both theorem and proof formalization with strong performance across the two tasks.

\bigskip


\section{Capabilities in Formal Verification for Open Problems}\label{sec:open}
In this section, we present Turing Agent's formal verification results on public mathematical research problems, demonstrating the agent's generalization capabilities.


In October 2026, OpenAI released an initial collection of 722 mathematical research manuscripts generated by an internal frontier model \citep{openai2026math}. We selected a subset of manuscripts that had not yet received complete formal verification and used Turing Agent to formalize and verify them.
We examine the formalization traces of three selected TCS problems concerning uniform matrix hitting for noncommutative formulas, hypergraph thresholds, and random 3-SAT hitting time. These traces document Turing Agent's problem analysis, proof decomposition, formal verification, and iterative refinement, as well as the construction and updates of the proof DAG throughout the process. The three cases are presented in order of increasing formalization difficulty, considering both the inherent mathematical complexity of the problems and the extent of existing formalization.

For brevity, the following three case studies provide concise accounts of Turing Agent's proof processes. More detailed analyses are provided in the Appendices~\ref{app:matrix}, \ref{app:hypergraph}, and \ref{app:3sat}.




\subsection{Uniform Matrix Hitting Points in Every Positive Characteristic}
Polynomial identity testing asks whether an arithmetic formula computes the zero polynomial. For noncommutative formulas, the order of multiplication matters, making identity testing more challenging. A central question is whether explicitly constructed matrices can detect all nonzero formulas within a given size bound through one fixed matrix substitution.

\begin{targetbox_blue}
\paragraph{Problem Definition.}
Given a prime $p$, a number of variables $n$, and a formula size bound $s$, the goal is to construct matrices $(T_1,\ldots,T_n)\in\operatorname{Mat}_d(\mathbb F_p)^n$ such that, for every field $F$ of characteristic $p$ and every nonzero polynomial $f$ computed by a division-free noncommutative arithmetic formula of size at most $s$ over $F$,
$f(T_1,\ldots,T_n)\neq 0$.
\end{targetbox_blue}

A noncommutative arithmetic formula consists of scalar and variable leaves combined through addition and ordered multiplication, with its size defined by the number of nodes. The same matrix tuple must detect all nonzero formulas within the size bound over arbitrary fields of characteristic $p$.

The full positive-characteristic result requires a deterministic construction over $\mathbb{F}_p$ with dimension $d=O(n^3s^6)$ in polynomial bit time, and also extends to acyclic algebraic branching programs (ABPs). The formalization examined here instead targets the corresponding characteristic-zero result, which uses an explicit rational matrix construction applicable to all characteristic-zero coefficient fields.

\begin{targetbox_yellow}
\paragraph{Target Theorem.}
For any positive integers $n,s$, there exists an explicitly constructed tuple of rational matrices $(T_1,\ldots,T_n)\in\operatorname{Mat}_d(\mathbb Q)^n$, where $d=1$ if $n=s=1$ and $d=2ns^2$ otherwise, such that for every field $F$ of characteristic zero and every nonzero polynomial $f$ computed by a division-free noncommutative arithmetic formula of size at most $s$ over $F$,
\[
f(T_1,\ldots,T_n)\neq 0.
\]
\end{targetbox_yellow}

\paragraph{Existing Formalization Provided by OpenAI.}
The existing characteristic-zero formalization provides an explicit rational matrix generator and a uniform hitting theorem for noncommutative arithmetic formulas, together with reusable results for formula representation and matrix evaluation. The mathematical argument involving power-series independence, differential systems, and vanishing-order bounds is encapsulated in the existing hitting theorem.

\paragraph{Turing Agent's Formal Proof Process.}
Turing Agent initially attempted to reuse existing results by invoking \texttt{complete\_result} and \texttt{universal\_hitting}, but encountered errors. After inspecting the library interfaces, it identified \texttt{completeClaim\_of\_hitting}, which reduces the target theorem to the uniform hitting property of the explicit rational matrix tuple. This allowed Turing Agent to focus on connecting the target formula representation to the existing hitting theorem.

Turing Agent first explored structural induction over four formula cases: scalars, variables, addition, and multiplication. It resolved the scalar and variable cases by checking nonzero matrix entries and proving the required dimension bounds, but the addition and multiplication cases remained unresolved. To organize the remaining proof, Turing Agent developed an eight-subgoal plan covering power-series independence, acyclic linear representations, differential systems, vanishing-order bounds, coefficient truncation, dimension estimates, special cases, and construction complexity.

The mathematical argument underlying the existing characteristic-zero hitting theorem associates each nonzero formula $f$ with a nonzero power series $h_f$. An acyclic linear representation with $m\leq 2s$ states yields an associated first-order differential system, whose multiplicity estimate gives
\[
\operatorname{ord}_0(h_f)
<(n-1)\binom{m}{2}+m
\leq 2ns^2.
\]
Together with the identity $f(T_1,\ldots,T_n)e_0=\pi_d(h_f)$, this bound establishes the required uniform nonzero matrix evaluation for the corresponding rational matrix construction. Rather than proving these components separately, Turing Agent returned to the existing \texttt{universal\_hitting} theorem, which encapsulates the argument and handles arbitrary formula structures.

The remaining challenge was connecting the formula representation in the target theorem to the implementation layer. Turing Agent used the recursive translation \texttt{toImplementation}, writing $\widehat f$ for the translated formula. Size and coefficient preservation establish $\operatorname{size}(\widehat f)=\operatorname{gates}(f)$ and $\operatorname{coeff}(\widehat f,w)=\operatorname{coefficient}(f,w)$, allowing the size constraint and nonzero coefficient condition to be transferred. Applying \texttt{universal\_hitting} then establishes nonzero matrix evaluation in the implementation layer, replacing the earlier, originally attempted but incomplete case-analysis approach.

Transferring this result back to the original representation introduced a dependent-type mismatch involving matrix dimensions and index types. Simplification and rewriting failed because dimensions determine matrix index types. Turing Agent used \texttt{convert} to expose the required equalities, applied \texttt{congrArg Fin} to align index types, and used equality elimination to resolve dependencies. Formula-evaluation preservation and matrix-entry consistency then completed the conversion.

Finally, Turing Agent established the target theorem, confirming the characteristic-zero uniform hitting property for the explicit rational matrix tuple and the resulting proof passed independent Lean kernel verification.

\subsection{A Uniform Influence Bound for Hypergraph Properties} 
Understanding when a random graph develops a property is a fundamental problem in random graph theory. A central question is how quickly the probability of this property changes as edges become more likely. This work extends the study of threshold phenomena from graphs to hypergraphs through the relationship between variance and total influence.

\begin{targetbox_blue}
\paragraph{Problem Definition.}
An $r$-uniform hypergraph has exactly $r$ vertices in each hyperedge. Consider a random $r$-uniform hypergraph on $n$ vertices, where each possible hyperedge is included independently with probability $p$. The objects of interest are Boolean hypergraph properties $f$ that are invariant under vertex permutations.
\end{targetbox_blue}

Let $\operatorname{Var}_p(f)$ denote the variance of $f$ under this model, and $I_p(f)$ its total influence, defined as the sum of the probabilities that flipping each possible hyperedge changes the value of $f$.

\begin{targetbox_yellow}
\paragraph{Target Theorem.}
For every integer $r\geq 3$, there exists a constant $C_r>0$, depending only on $r$, such that for all $n\geq r$, $p\in(0,1)$, and vertex-permutation-invariant Boolean hypergraph properties $f$,
\[
\operatorname{Var}_p(f)
\leq
\frac{C_r}{(\log n)^{r/(r-1)}} I_p(f).
\]
\end{targetbox_yellow}

The bound holds without any monotonicity assumption. Here, $\log$ denotes the natural logarithm, and $I_p(f)$ does not include a factor of $p(1-p)$.

\paragraph{Existing Formalization Provided by OpenAI.}
The existing formalization establishes a variance-influence bound $\operatorname{Var}_p(f)\leq C I_p(f)$ and a threshold-width result for monotone graph properties, along with reusable tools for Fourier analysis, influence calculations, coordinate restrictions, and averaging. For $r$-uniform hypergraphs with $r\geq 3$, it provides model definitions, auxiliary lemmas, and a framework for the target theorem, but leaves a parameter-selection lemma unfinished.

\begin{figure}[!h]
    \centering
    \includegraphics[width=\linewidth]{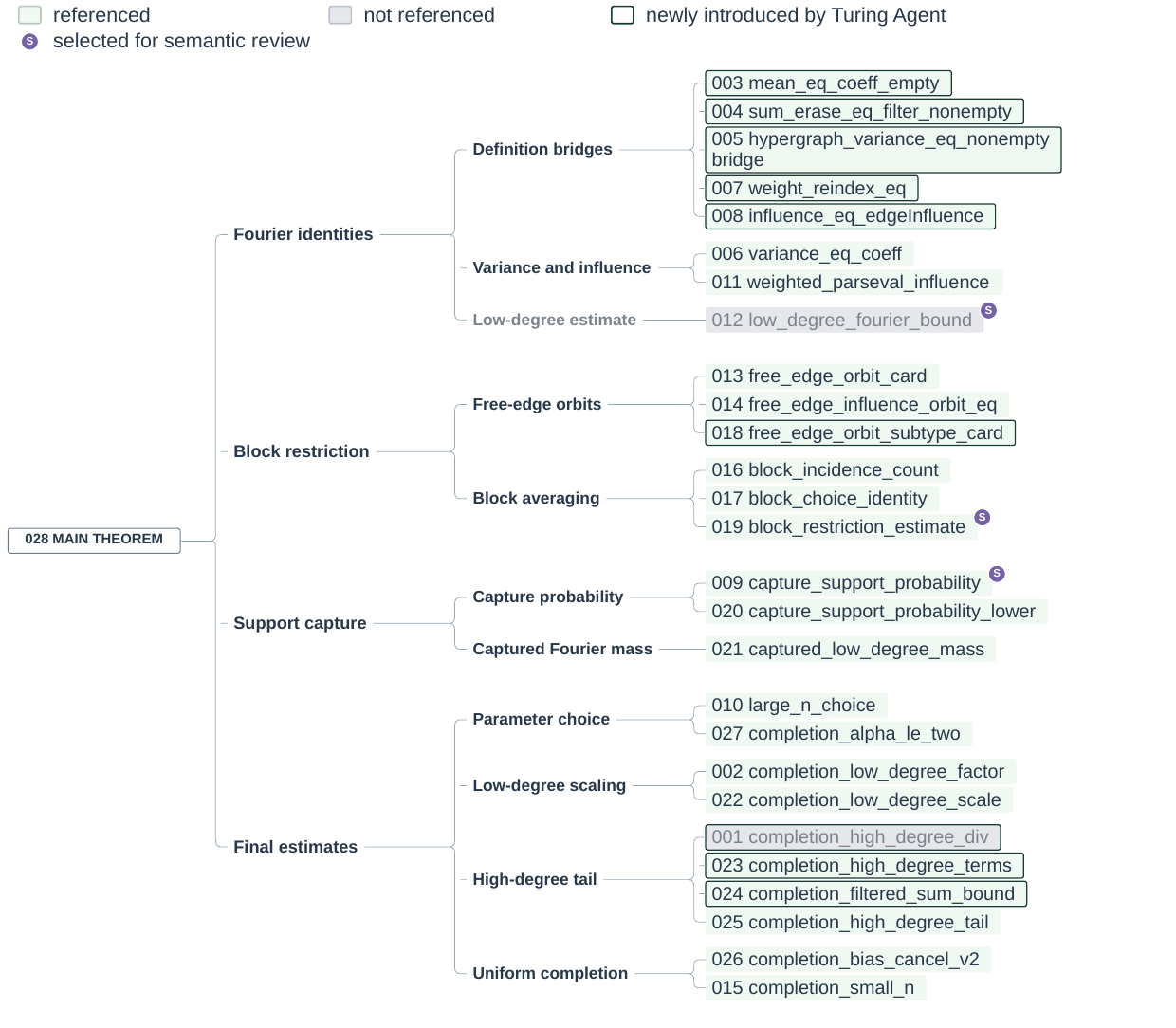}
    \caption{A simplified representation of the proof DAG for Problem \textit{A uniform influence bound for hypergraph properties}. The relationships between nodes are omitted in the figure. Each leaf node represents a lemma or subtheorem, while the intermediate nodes indicate logical categories rather than actual nodes in the DAG. Green terminal nodes indicate those actually used in the proof, while gray nodes indicate those not used in the final proof. Leaf nodes with bold borders represent lemmas or theorems introduced by Turing Agent during the proving process and not present in the original paper. Purple \textbf{S} markers indicate nodes that are selected for semantic review.}
    \label{openai-186-01-dag}
\end{figure}

\paragraph{Turing Agent's Formal Proof Process.}
Turing Agent first connects the hypergraph definitions to Fourier analysis. Directly expanding the variance and rewriting nested sums leads to matching failures and type-checking timeouts, so it introduces separate lemmas for the mean correspondence and summation-index conversions. Writing $\widehat f_p(S)$ for the Fourier coefficient indexed by a set $S$ of hyperedges, it establishes two identities underlying the low-degree and high-degree estimates:
\[
\operatorname{Var}_p(f)
=
\sum_{S\ne\varnothing}\widehat f_p(S)^2,
\qquad
\sum_S |S|\widehat f_p(S)^2
=
p(1-p)I_p(f).
\]

For the low-degree estimate, Turing Agent selects a vertex block $B$ of size $m$, retains hyperedges intersecting $B$ as free coordinates, and fixes the remaining coordinates to construct $f_{B,y}$. It proves that vertex permutations within $B$ preserve the restriction and that hyperedges in the same orbit have equal influence. Since each free hyperedge belongs to an orbit of at least $m$ edges when $m\geq r+1$, it obtains $I_{p,e}(f_{B,y})\leq I_p(f_{B,y})/m$. However, the existing orbit-size lemma counts all hyperedges, while the influence bound applies only to the subtype of free hyperedges. Turing Agent resolves this type mismatch by introducing an auxiliary lemma connecting the two counting representations.

Turing Agent then establishes averaging identities by exchanging summations over vertex blocks and hyperedges, and uses support capture to relate the low-degree Fourier terms of $f$ to those of its restrictions. It introduces the parameters $\alpha=r/(r-1)$, $k=\sqrt{p(1-p)}\log m/64$, and $L=(k/r)^\alpha$, subject to the capture condition $rL(m-1)/(n-1)\leq 1/2$. Combining the averaging upper bound with the capture-probability lower bound yields
\[
\sum_{1\leq |S|\leq L}\widehat f_p(S)^2
\leq 2r m^{-1/4}I_p(f).
\]

For the high-degree estimate, directly handling the Fourier summation again causes type-checking timeouts. Turing Agent instead proves a general inequality for $L>0$ and nonnegative sequences $a_i,b_i$ over an arbitrary finite index type:
\[
\sum_{a_i>L}b_i
\leq
\frac{1}{L}\sum_i a_i b_i.
\]
Applying this inequality with $a_S=|S|$ and $b_S=\widehat f_p(S)^2$, together with the weighted Fourier identity, gives
\[
\sum_{|S|>L}\widehat f_p(S)^2
\leq
\frac{p(1-p)}{L}I_p(f).
\]

Finally, Turing Agent proves that $m=\lfloor\sqrt n\rfloor$ satisfies the parameter conditions for sufficiently large $n$ and combines the two estimates. The remaining small-scale cases follow from $\operatorname{Var}_p(f)\leq I_p(f)$, completing the target theorem. The process completes 27 auxiliary proof nodes, and the final result passes export compilation and independent Lean kernel checking. Figure~\ref{openai-186-01-dag} illustrates a simplified representation of the theorems, lemmas, and other results formally verified by Turing Agent for this problem, with relationships between nodes omitted.

\subsection{Linear Variance of the Random 3-SAT Hitting Time}
In random satisfiability problems, adding clauses can turn a satisfiable formula into an unsatisfiable one. A central question is how much the timing of this transition varies across random instances. Understanding these fluctuations helps determine how tightly the transition is concentrated.

\begin{targetbox_blue}
\paragraph{Problem Definition.}
A 3-SAT formula is a conjunction of clauses, each consisting of a disjunction of three literals on distinct Boolean variables. Over $n\geq 3$ variables, there are $8\binom{n}{3}$ possible clauses. Consider a random process in which clauses $C_1,C_2,\ldots$ are sampled independently and uniformly, with repetitions allowed. Let $F_m=C_1\land\cdots\land C_m$, with $F_0$ being the satisfiable empty formula. The first unsatisfiability time is defined as $H_n=\min\{m\geq 1:F_m\text{ is unsatisfiable}\}$.
\end{targetbox_blue}

The problem is to bound the variance $\operatorname{Var}(H_n)$ of this transition time. Previous work established an $O(n\log n)$ upper bound on $\operatorname{Var}(H_n)$. The goal is to improve this bound to $O(n)$.

\begin{targetbox_yellow}
\paragraph{Target Theorem.}
There exists a constant $C>0$, independent of $n$, such that for all $n\geq 3$, $\operatorname{Var}(H_n)\leq Cn$.
\end{targetbox_yellow}

Combined with Wilson's separation theorem, this yields $\operatorname{Var}(H_n)=\Theta(n)$. The formalization targets the linear upper bound and the foundational properties of the random model.

\paragraph{Existing Formalization by OpenAI.}
The existing formalization defines the random $k$-SAT model and its first unsatisfiability time $H_{n,k}$, establishing $\operatorname{Var}(H_{n,3})=O(n\log n)$ and $\operatorname{Var}(H_{n,k})=\Theta(n)$ for fixed $k\geq 4$. It provides coordinate-deletion inequalities, collision-case estimates, exponential tail bounds, and truncation-removal lemmas. However, substantial proof components required to establish the linear variance upper bound for $k=3$ remained unformalized.

\paragraph{Turing Agent's Formal Proof Process.}
Turing Agent first establishes foundational results on clause counting, survival probabilities, moment identities, and small-instance bounds, together with an explicit truncation-error estimate. Using OpenAI's existing coordinate-deletion inequality, it reduces the main theorem to bounding squared deletion delays for $T=\min(H_n,10n)$, where $D_i$ denotes the increase in the capped hitting time caused by omitting clause $C_i$ without renumbering subsequent clauses. A uniform bound $\mathbb{E}[D_i^2]\leq C$ then implies $\operatorname{Var}(T)\leq 10Cn$.

To remove the logarithmic factor in deletion-delay estimates, Turing Agent shifts from survival-time estimates for assignment sets to potential-function drift along the clause process. It introduces $P_d(S)=\sum_{j<d}q_j(S)$, where $q_j(S)$ is the probability that $j$ independent random 2-clauses eliminate all assignments in $S$. Expanding nested averages causes rewriting failures, so Turing Agent proves the filtering-exchange and averaging identities separately. These results enable the drift estimate and an $O(d^{3/2})$ cumulative bound through telescoping sums.

Turing Agent then extends the cumulative estimate to processes involving deterministic filtering steps by proving the monotonicity of the potential function and controlling the additional cost through the root-incidence count $Z$. Combining this estimate with the remaining survival-time bound and three independent root-variable tests yields
\[
2\bigl(C_0d^{3/2}+Z\bigr)
\bigl(23d^{3/2}+Z\bigr)
=O(d^3),
\qquad d=\max(1,Z).
\]
This establishes the required deletion-delay estimate for the collision-free case.

For the remaining collision cases, Turing Agent combines the two-root-test estimate for single collisions and the truncation bound for multiple collisions with auxiliary lemmas for clause decomposition, resampling, and averaging over deleted coordinates. It further establishes exponential moment bounds for root-incidence counts, factorial moment bounds for collision counts, and mixed-moment estimates. Averaging over the three collision cases under the original clause distribution yields $\mathbb{E}[D_i^2]=O(1)$, and hence $\operatorname{Var}(T)=O(n)$.

Finally, Turing Agent applies the truncation-removal lemma $\operatorname{Var}(H_n)\le 2\operatorname{Var}(T)+O(1)$ to extend the bound to the original hitting time. To connect the random clause sequence model with the finite-counting representation in the formalization, it establishes satisfiability-preserving bijections between clauses and clause tuples and relates moments through tail-sum and integral identities. With the small-instance bounds, these results establish $\operatorname{Var}(H_n)\le Cn$ for all $n\ge 3$. The proof contains 108 verified auxiliary nodes, and the root theorem passes export compilation and Lean kernel verification.

\section{Conclusion and Outlook}\label{sec:conclusion}

In this work, we presented \textbf{Turing Agent}, a unified harness for faithful formal verification in theoretical computer science. The system adopts a two-scale design that coordinates theorem formalization and proof formalization, while incorporating structured proof decomposition, DAG-based dependency management, parallel solving, an experience pool, and selective semantic review. This design enables Turing Agent to address not only formal proof search, but also the broader challenge of maintaining semantic fidelity between an informal research claim, its formal statement, and the resulting machine-checked proof.

Our experiments show that Turing Agent outperforms all evaluated baselines on Formal TCS Bench with both models and achieves competitive performance on RLMEval. In particular, the results on Formal TCS Bench highlight the importance of evaluating semantic faithfulness in addition to compilation success, while the strong performance on RLMEval demonstrates the effectiveness of the harness for research-level Lean proof construction. Beyond standard benchmarks, Turing Agent successfully completes the formal verification of several research-level TCS problems with incomplete formal developments, including cases involving long dependency chains and large numbers of auxiliary claims.

These results suggest that reliable AI-assisted research in theoretical computer science requires more than stronger base models alone. A verification-oriented harness that explicitly coordinates semantic alignment, formal proof construction, and long-horizon reasoning can substantially improve the reliability and scalability of automated formal reasoning. We view Turing Agent as a step toward AI systems that can not only propose sophisticated theoretical results, but also produce trustworthy, machine-checkable evidence for them. Future work will focus on extending the system to broader areas of theoretical computer science, improving automated semantic checking and proof decomposition, and more tightly integrating formal verification with research-level problem discovery.

\FloatBarrier
\bibliographystyle{plainnat}
\bibliography{reference}


\clearpage
\appendix
\crefalias{section}{appendix}
\crefalias{subsection}{subappendix}
\crefalias{subsubsection}{subsubappendix}

\section{Detailed Verification Process for Problem {\em Uniform Matrix Hitting Points in Every Positive Characteristic}}
\label{app:matrix}
This section provides additional details of Turing Agent's formal verification of the characteristic-zero companion result associated with the problem \textit{Uniform Matrix Hitting Points in Every Positive Characteristic}.

\paragraph{Additional Details on OpenAI's Existing Formalization}
The existing OpenAI formalization library provides the target specification, the matrix generation procedure and its correctness guarantees, the mathematical hitting theorem, and interfaces connecting different formula representations. The specification describes formulas using node counts, word coefficients, and matrix evaluation, whereas the mathematical implementation uses a separate recursive formula representation to organize coefficient operations, linear representations, and power-series arguments.

The generator defines explicit matrix entries. For $1\le i\le n$ and $0\le r,q<d$, the general construction uses
\[
(T_i)_{rq}=
\begin{cases}
\displaystyle\frac{(-1)^{r-q-1}}{r\,i^{r-q}}, & q<r,\\
0, & q\ge r.
\end{cases}
\]
The special case $n=s=1$ uses $T_1=[1]$. Existing results establish the output length, positive denominators, encoding bounds, exact decoding, and generation cost. The \texttt{completeClaim\_of\_hitting} interface combines these guarantees, reducing the complete target theorem to the uniform hitting property at the specification level.

The mathematical hitting property is already established by \texttt{universal\_hitting}. Given a nonzero word coefficient of a formula over a characteristic-zero field, this theorem proves that evaluation at the generated matrices is nonzero. The underlying proof uses word-based power-series encodings, acyclic linear representations of formulas, multiplicity bounds for differential systems, and the correspondence between truncated power series and matrix evaluation. The library also handles the special case $n=s=1$.

Existing translation results further connect the two formula representations. The recursive translation \texttt{toImplementation} preserves formula size, every word coefficient, and evaluation under arbitrary matrix substitutions. Additional generator interfaces establish the equality of matrix dimensions and entries across the two representations. These results provide the mathematical and computational guarantees required by the target theorem, leaving their composition under the target's quantifiers and dependent types as the remaining formalization task.

\paragraph{Overview of the Proof.}
The mathematical proof encodes a nonzero noncommutative polynomial as a nonzero formal power series, then uses the formula size to bound the order of vanishing. A finite truncation consequently retains a nonzero coefficient, while explicit matrices realize the corresponding truncated integration operators. Existing generator guarantees complete the construction and complexity claims.

For words over the alphabet $\{1,\ldots,n\}$, define
\[
h_{\varnothing}=1,\qquad
h_{iw}(z)=\int_0^z\frac{h_w(t)}{t+i}\,\mathrm{d}t.
\]
The integration is defined coefficientwise on formal power series. Characteristic zero ensures the required integer inverses and distinct pole locations. An existing independence theorem establishes the linear independence of these word series, so a nonzero polynomial $f=\sum_w f_wx_w$ induces a nonzero series $h_f=\sum_w f_wh_w$.

The required truncation length follows from a small linear representation of the formula. A formula of size at most $s$ admits an acyclic representation with at most $m\le2s$ states, yielding constant vectors $u,v$, matrices $B_i$, and a fundamental matrix $G$ satisfying
\[
h_f=uGv,\qquad
G'=\left(\sum_{i=1}^n\frac{B_i}{z+i}\right)G,
\qquad G(0)=I.
\]
Setting $q(z)=\prod_{i=1}^n(z+i)$ transforms the system into $qG'=PG$, where $q(0)\neq0$ and the entries of $P$ have degree at most $n-1$. An existing Wronskian multiplicity estimate then gives
\[
\operatorname{ord}_0(h_f)
<(n-1)\binom{m}{2}+m
\le 2ns^2.
\]
Therefore, at least one of the first $2ns^2$ coefficients of $h_f$ is nonzero. Although the differential system depends on the formula, the truncation length depends only on $n$ and $s$, making a uniform hitting construction possible.

Let $\pi_d$ retain the coefficients of degrees $0,\ldots,d-1$ as a column vector. The explicit matrices $T_i$ implement the operators $a\mapsto\int_0^z a(t)/(t+i)\,\mathrm{d}t$ on this truncated space. Extending the correspondence from words to formulas gives
\[
f(T_1,\ldots,T_n)e_0=\pi_d(h_f),
\qquad e_0=\pi_d(1).
\]
For $d=2ns^2$, the multiplicity bound makes the right-hand side nonzero, establishing the hitting property. When $n=s=1$, the formula size restriction leaves only scalar or single-variable formulas, handled by $T_1=[1]$. Finally, the translation preserves formula size, coefficients, and evaluation, allowing the mathematical hitting theorem to be combined with the generator's output and complexity guarantees.

\paragraph{Turing Agent’s Formal Proof Process.}
Turing Agent began by reusing existing library results. Initial attempts to invoke \texttt{complete\_result} and \texttt{universal\_hitting} directly failed because the identifiers were not visible in the current proof environment. After inspecting the available interfaces, Turing Agent identified \texttt{completeClaim\_of\_hitting} and introduced the parameters $n,s$ together with their positivity assumptions. This reduced the complete target to the specification-level uniform hitting property, incorporating the existing guarantees for generator output, encoding, and computational cost.

\begin{figure}[!htbp]
    \centering
    \includegraphics[width=\linewidth]{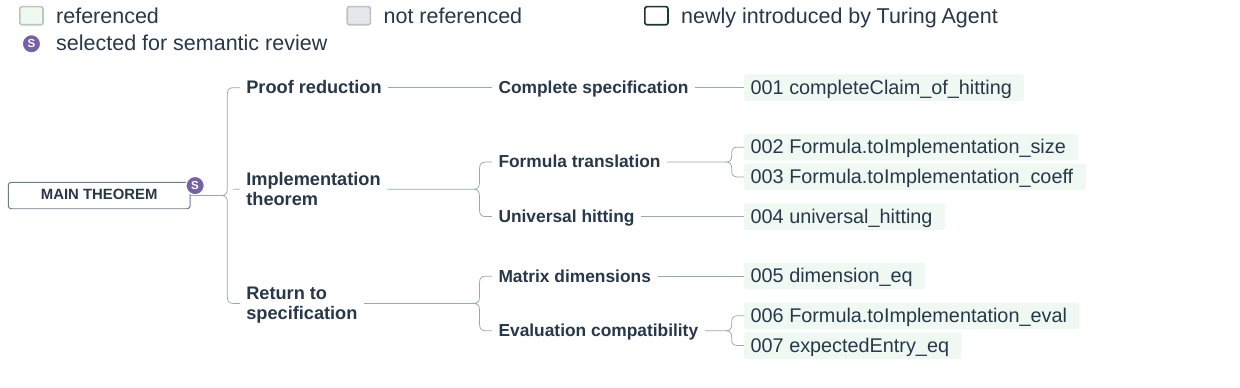}
    \caption{A simplified representation of the proof DAG for Problem {\em Uniform Matrix Hitting Points in Every Positive Characteristic}.}
    \label{openai-116-01-dag}
\end{figure}

Turing Agent next explored a proof by structural induction on formulas, considering scalar, variable, addition, and multiplication cases. The scalar case used a matrix coordinate in positive dimension to establish nonvanishing. For variables, Turing Agent separated the one-dimensional special case from the general construction, using the matrix entry $(T_i)_{1,0}=1/i\neq0$ in the latter. Establishing the validity of this coordinate required proving that the dimension exceeds $1$ and switching from linear arithmetic to reasoning involving multiplication. These local attempts resolved the scalar and variable cases, while the addition and multiplication cases remained open.

To organize the remaining mathematics, Turing Agent proposed eight auxiliary goals covering power-series independence, acyclic linear representations, differential systems, multiplicity bounds, coefficient truncation, dimension estimates, the special case, and construction complexity. Further inspection revealed that the existing \texttt{universal\_hitting} theorem already combined these dependencies. Specifically, the theorem establishes $h_f\neq0$ from a nonzero word coefficient, derives $\operatorname{ord}_0(h_f)<2ns^2$ using a system with at most $2s$ states, and obtains nonvanishing matrix evaluation through the identity relating truncated series to the first column of matrix evaluation. Turing Agent therefore reused this established result rather than proving the eight planned goals independently. None of these goals entered the final proof as newly established auxiliary theorems.

The remaining substantive task was to connect the target formula representation with the library implementation. For a target formula $f$, Turing Agent applied \texttt{toImplementation} to obtain an implementation-level formula $\widehat{f}$. Size preservation converted $\operatorname{gates}(f)\le s$ into $\operatorname{size}(\widehat{f})\le s$, while coefficient preservation transferred the same nonzero word coefficient witness $w$, giving $\operatorname{coeff}(\widehat{f},w)\neq0$. These conditions matched the hypotheses of \texttt{universal\_hitting}, yielding nonzero evaluation of $\widehat{f}$ at the generated matrices.

Transferring this result back to the target specification required additional work with dependent types. Direct simplification and rewriting of matrix dimensions produced type mismatches because matrix indices have type \texttt{Fin d}, which depends on the dimension expression. Turing Agent used \texttt{convert} to separate the required equalities, applied \texttt{congrArg Fin} to transport dimension equality to equality of index types, and eliminated the resulting equalities. The established matrix-entry correspondence and evaluation-preservation theorem then connected the two matrix evaluations. This completed the transfer from implementation-level nonvanishing to the target conclusion $f(T_1,\ldots,T_n)\neq0$.

Finally, Turing Agent assembled the formula translation, existing hitting theorem, and generator completeness interface into \texttt{ProofTarget\_116}. The resulting proof preserved arbitrary positive integers $n,s$, arbitrary characteristic-zero coefficient fields, and the one-dimensional special case, while incorporating the output encoding, decoding, and computational cost guarantees. The exported proof passed Lean verification with formal status \texttt{export\_verified}, and the final fidelity review returned \texttt{accepted}, confirming consistency with the target statement. The completed formalization thus establishes a verified connection between the existing mathematical results and the complete theorem specification. Figure~\ref{openai-116-01-dag} illustrates a simplified representation of the proof DAG for this problem, with relationships between nodes omitted.

\section{Detailed Proof Process for Problem {\em A uniform influence bound for hypergraph properties}}
\label{app:hypergraph}
This section provides additional details of Turing Agent's proof processes for the problem \textit{A uniform influence bound for hypergraph properties}.

\paragraph{Additional Details on OpenAI's Existing Formalization} The existing formalization covers the graph case, including variance-influence inequalities for graph properties and threshold-width results for monotone graph properties. It also provides reusable tools for Fourier analysis, influence calculations, coordinate restrictions, and averaging.

For hypergraphs, the code already contains the model definitions, auxiliary lemmas, and a framework for the main theorem. One parameter-selection lemma, however, remains unfinished. Turing Agent builds on this existing code, reuses the available results, and completes the missing proof steps needed to establish the hypergraph bound.
\paragraph{Overview of the Proof.}
The formal proof can be divided into the following four parts according to its dependency structure. The low-degree and high-degree estimates form two separate branches built on the same variance decomposition.

\begin{enumerate}
\item \textbf{Variance decomposition.} Connect the variance and influence of the hypergraph property to the existing Fourier analysis tools. Establish that the variance equals the sum of squares of the nonempty Fourier coefficients, while the sum of their squares weighted by their degrees equals $p(1-p)I_p(f)$. This reduces the target theorem to separately estimating the low-degree and high-degree contributions.

\item \textbf{Low-degree estimate.} Select a block of vertices, retain the hyperedges intersecting this block as free coordinates, and fix only the hyperedges outside the block. Using vertex-permutation invariance, show that the restricted function has sufficiently large coordinate orbits, allowing the existing low-degree estimate to be applied and averaged. Meanwhile, use low-degree vertices in the support of the hypergraph to establish a lower bound on the probability that a low-degree Fourier term of the original function remains a nonempty low-degree term under a random restriction. Combining this probability lower bound with the averaging upper bound controls the low-degree contribution of the original function.

\item \textbf{High-degree estimate.} Use the weighted sum-of-squares identity to directly bound the high-degree tail in terms of total influence. The truncation parameter is chosen to obtain the required logarithmic decay. Since $r\geq 3$, the factor depending on $p$ can be bounded uniformly, leaving a final constant that depends only on $r$.

\item \textbf{Parameter existence and combination of estimates.} Prove that, when the number of vertices is sufficiently large, choosing the block size as $\lfloor\sqrt n\rfloor$ satisfies the conditions required by both estimates. Combine the low-degree and high-degree upper bounds. For the remaining values of $n$, use a general variance-influence inequality to complete the proof. After choosing a common constant for both cases, obtain the target inequality for all $n\geq r$ and $p\in(0,1)$, without requiring a monotonicity assumption.
\end{enumerate}

\paragraph{Turing Agent's Formal Proof Process.}
Turing Agent begins by connecting the hypergraph definitions to the Fourier analysis interfaces, progressively establishing the proof dependencies required by the target theorem. Initially, it attempts to expand the variance, mean, and weights directly and perform rewrites within nested sums. This leads to repeated matching failures and type-checking timeouts. It subsequently abstracts the correspondence involving the mean and the conversion of summation indices into separate lemmas, then uses these lemmas to assemble the variance interface. Writing $\widehat f_p(S)$ for the Fourier coefficient indexed by a set $S$ of hyperedges, it establishes two key identities which provide a common foundation for the subsequent estimates of low-degree and high-degree Fourier coefficients:

\[
\operatorname{Var}_p(f)
=
\sum_{S\ne\varnothing}\widehat f_p(S)^2,
\qquad
\sum_S |S|\widehat f_p(S)^2
=
p(1-p)I_p(f).
\]

For a vertex block $B$ of size $m$, Turing Agent retains the hyperedges intersecting $B$ as free coordinates, fixes all remaining coordinates to an assignment $y$, and constructs the restricted function $f_{B,y}$. It then considers permutations of the vertices within $B$ and proves that the induced permutations of hyperedges preserve the free coordinates and fix all coordinates outside $B$. Since the restricted function is invariant under these permutations, edges belonging to the same orbit have equal influence. In particular, for the orbit $\mathcal O_B(e)$ of a free hyperedge $e$, whose members agree with $e$ outside $B$, the orbit contains at least $m$ edges when $m\geq r+1$. Since these edges all have the same influence, their combined contribution to the total influence is at least $m I_{p,e}(f_{B,y})$. Consequently, $I_{p,e}(f_{B,y})\leq I_p(f_{B,y})/m$.

During this process, the orbit-size lemma counts all hyperedges, whereas the influence bound applies to the subtype consisting of free hyperedges. The difference in their types prevents the proof from being completed through direct lemma application, simplification, or numerical type conversion. Turing Agent subsequently introduces an auxiliary lemma connecting the two counting representations, resolving this step.

For the averaging argument, Turing Agent exchanges the order of summation over vertex blocks and hyperedges, reducing the problem to counting how many vertex blocks a given edge intersects. It proves the corresponding counting identities and combinatorial coefficient identities separately.

The support-capture argument then relates the low-degree Fourier terms of the original function to those of the restricted function. To establish this relation, Turing Agent introduces the parameters
\[
\alpha=\frac{r}{r-1},\qquad
k=\frac{\sqrt{p(1-p)}\log m}{64},\qquad
L=\left(\frac{k}{r}\right)^\alpha,
\]
which are chosen to satisfy the capture condition
\[
\frac{rL(m-1)}{n-1}\leq 1/2.
\]

Under this condition, the averaging upper bound can be combined with the lower bound on the capture probability to obtain the low-degree estimate
\[
\sum_{1\leq |S|\leq L}\widehat f_p(S)^2
\leq 2r m^{-1/4}I_p(f).
\]

The construction of vertex permutations, orbit influence bounds, averaging identities, and support capture together form the main chain of auxiliary lemmas for the low-degree estimate.

For the high-degree estimate, Turing Agent takes a different approach. Directly handling the large sum involving hyperedge types and Fourier coefficients repeatedly leads to type-checking timeouts, while an attempt to apply an existing division lemma also fails. Turing Agent therefore first proves a general summation inequality over an arbitrary finite index type. Specifically, for $L>0$ and nonnegative sequences $a_i$ and $b_i$, it establishes that
\[
\sum_{a_i>L}b_i
\leq
\frac{1}{L}\sum_i a_i b_i.
\]

It then applies this inequality with $a_S=|S|$ and $b_S=\widehat f_p(S)^2$. Together with the weighted Fourier identity established earlier, this gives the high-degree estimate
\[
\sum_{|S|>L}\widehat f_p(S)^2
\leq
\frac{p(1-p)}{L}I_p(f).
\]

By reducing the original Fourier summation to a general finite-sum inequality, this approach avoids the difficulties encountered in the direct proof.

After completing the low-degree and high-degree estimates, Turing Agent shows that, for sufficiently large $n$, the choice
$m=\lfloor\sqrt n\rfloor$
satisfies all the parameter conditions established earlier. Combining the two estimates gives the desired bound for sufficiently large $n$. For the remaining small-scale cases, Turing Agent uses the elementary inequality
$\operatorname{Var}_p(f)\leq I_p(f)$,
thereby completing the proof of the target theorem.

The process establishes and completes 27 auxiliary proof nodes and the final result passes export compilation and independent Lean kernel checking.

The exploration also produces two auxiliary results: a low-degree estimate packaged for the full hypergraph coordinate space, and a lemma converting a product bound into a division bound. The final proof instead uses a general library tool applicable to free coordinates in the restriction argument and a general summation inequality for the high-degree estimate.

\section{Detailed Proof Process for Problem {\em Linear Variance of the Random 3-SAT Hitting Time}}
\label{app:3sat}
This section provides additional details of Turing Agent's proof processes for the problem \textit{Linear Variance of the Random 3-SAT Hitting Time}.

\paragraph{Additional Details on OpenAI's Existing Formalization} OpenAI's existing formalization establishes a general random $k$-SAT clause model, an infinite sequence of independently and uniformly sampled clauses, and the first unsatisfiability time together with its truncated version. Let $H_{n,k}$ denote the first unsatisfiability time in this random model. For every fixed $k\ge 3$, the formalized results establish the existence of a constant $C_k$, depending only on $k$, such that
\[
    \operatorname{Var}(H_{n,k})\le C_k n\ell_k(n),
    \qquad \forall n\ge k,
\]
where
\[
    \ell_k(n)=
    \begin{cases}
        1+\log n, & k=3,\\
        1, & k\ge 4.
    \end{cases}
\]
In addition, OpenAI proved a linear variance lower bound for sufficiently large $n$. Thus, the existing formalization establishes a $\Theta(n)$ variance bound for $k\ge 4$, while the upper bound for $k=3$ still contains a logarithmic factor.

Regarding the proof techniques, OpenAI formalized a coordinate-deletion inequality that bounds the variance of the truncated hitting time by the sum of the second moments of individual clause-deletion delays. For the variables involved in a deleted clause, its formalization also establishes independent tests, a classification of collision cases, and estimates of the remaining survival time. Furthermore, exponential tail bounds and truncation-removal lemmas allow the variance bound for the truncated hitting time to be extended to the original hitting time. Together, these results provide a framework for deriving global variance bounds from local deletion-delay estimates.

For $k=3$, improving the variance upper bound to $O(n)$ still requires bounds on the increments and cumulative behavior of a potential function. Combined with average bounds over collision cases, these estimates are needed to control the expected squared deletion delay at each position by $O(1)$. In addition, since the target theorem defines survival probabilities and moments through finite counting, their correspondence with the existing formalized random clause model and its variance must also be established.

\paragraph{Overview of the Proof.}
  The proof reduces the variance bound for the first unsatisfiability time to second-moment estimates for individual clause-deletion delays. A uniform bound is then established through potential-function estimates, independent root-variable tests, and collision
  analysis.

  \begin{enumerate}
      \item \textbf{Reduction to deletion-delay estimates.}
      Satisfiability-preserving bijections and moment identities connect the finite-counting definitions with the random clause sequence model. For $n\ge 6$, set $T=\min(H_n,10n)$, and let $D_i$ denote the truncated delay caused by deleting the $i$-th clause. The
      coordinate-deletion inequality gives
      \[
          \operatorname{Var}(T)
          \le \sum_{i=1}^{10n}\mathbb{E}[D_i^2].
      \]
      Thus, a uniform second-moment bound implies a linear variance bound for $T$.

      \item \textbf{Uniform bound on squared deletion delays.}
      Fix the three variables of the deleted clause and let $d$ denote the larger of one and the number of other clauses involving these variables. A potential function based on the probability that random binary clauses eliminate a set of assignments yields a
      cumulative bound of order $O(d^{3/2})$ through its boundedness and drift properties. Combined with three independent root-variable tests and survival-time estimates, this gives an $O(d^3)$ conditional bound on the squared deletion delay in the collision-
      free case.

      Single and multiple collisions are controlled using two root-variable tests and the truncation bound, respectively. Moment bounds on root-incidence and collision counts then yield $\mathbb{E}[D_i^2]=O(1)$ uniformly.

      \item \textbf{Truncation removal.}
      A truncation-removal lemma extends the linear variance bound from $T$ to $H_n$. The cases $n=3,4,5$ are handled using geometric tail bounds.
  \end{enumerate}

  The key is that the potential-function estimate matches the cubic probability bound from the three independent tests, removing the logarithmic factor in the previous analysis.

\paragraph{Turing Agent's Formal Proof Process.}
Turing Agent first established the foundational results on probabilities and moments before reducing the proof of the linear variance bound to a uniform estimate on deletion delays. The initial work included clause counting, survival probabilities, moment tail-sum identities, second-moment bounds for small instances, and an explicit truncation-error bound of $120[2(7/8)^{10}]^n$. Turing Agent then used OpenAI's existing coordinate-deletion inequality to establish an auxiliary theorem: for $T=\min(H_n,10n)$, a uniform bound $\mathbb{E}[D_i^2]\le C$ for all positions implies $\operatorname{Var}(T)\le 10Cn$. This established the main objective for the subsequent proof.

To eliminate the logarithmic factor in existing deletion-delay estimates, Turing Agent shifted from survival-time estimates for individual assignment sets to potential-function drift along the process. It first specialized a survival-time product bound from OpenAI's existing formalization to 3-SAT and attempted to control the cumulative quantity. Further examination showed that estimates for individual sets still needed to be connected to the evolution of the process, while the threshold-integration method in the library reached a critical exponent and retained a logarithmic factor. Turing Agent therefore introduced the potential function $P_d(S)=\sum_{j<d}q_j(S)$ and established probability-increment comparisons, single-clause averaging identities, and drift lemmas. Direct expansion of nested averages initially caused rewriting failures. After proving the filtering-exchange and averaging identities separately, Turing Agent completed the drift estimate and obtained an $O(d^{3/2})$ cumulative bound through telescoping sums.

After establishing the cumulative estimate, Turing Agent extended it to the process required by the root-variable tests and used it to bound squared deletion delays in the collision-free case. It first proved the monotonicity of the potential function, extending the result from ordinary random clause processes to processes interleaved with deterministic filtering steps, whose additional cost was controlled by the root-incidence count $Z$. Turing Agent then combined the cumulative bound with the remaining survival-time estimate and three independent root-variable tests, obtaining the following upper bound on the squared reward after resampling:
\[
    2\bigl(C_0d^{3/2}+Z\bigr)
    \bigl(23d^{3/2}+Z\bigr)
    =O(d^3),
    \qquad d=\max(1,Z).
\]
This result was subsequently used to bound squared deletion delays in the collision-free case, completing the connection between the potential-function estimate and the root-variable tests.

After completing the collision-free estimate, Turing Agent combined the three conditional estimates under the original clause distribution to establish a uniform second-moment bound. OpenAI's existing formalization provided the two-root-test estimate for single collisions and the truncation bound for multiple collisions. Turing Agent then proved auxiliary lemmas for clause decomposition, preservation of averages under resampling, and averaging over the deleted coordinate. It also completed exponential moment bounds for root-incidence counts, factorial moment bounds for collision counts, and mixed-moment estimates. These results combined the three contributions into $\mathbb{E}[D_i^2]=O(1)$, which, together with the earlier variance reduction, yielded a linear variance bound for the truncated hitting time.

Finally, Turing Agent extended the linear variance bound from the truncated hitting time to the original hitting time using the existing truncation-removal lemma, which gives $\operatorname{Var}(H_n)\le 2\operatorname{Var}(T)+O(1)$. The explicit truncation-error bound proved earlier was retained as an auxiliary result. To connect this bound with the finite-counting variance in the target theorem, Turing Agent established satisfiability-preserving bijections between clauses and clause tuples, then used the equivalence between tail sums and integrals to relate the first and second moments in the two representations. 

\begin{figure}[htbp]
    \centering
    \includegraphics[width=0.95\linewidth]{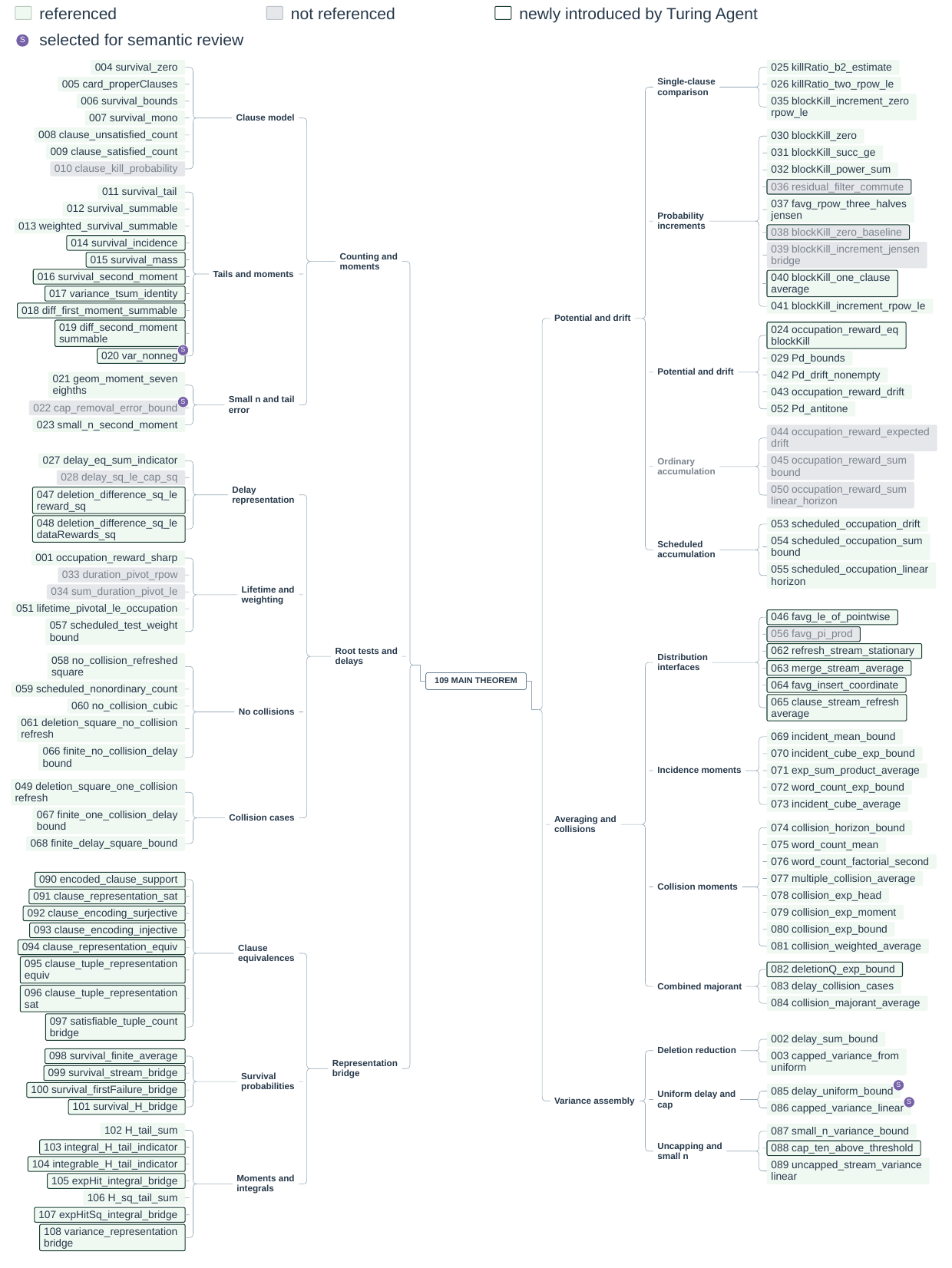}
    \caption{A simplified representation of the proof DAG for Problem {\em Linear Variance of the Random 3-SAT Hitting Time}.}
    \label{openai-235-02-dag}
\end{figure}

Combining these results with the small-instance estimates, Turing Agent established the existence of a constant $C>0$, independent of $n$, such that $\operatorname{Var}(H_n)\le Cn$ for all $n\ge 3$. The final proof involved 108 verified auxiliary nodes, and the root theorem passed export compilation and Lean kernel verification. Figure~\ref{openai-235-02-dag} illustrates a simplified representation of the proof DAG for this problem, with relationships between nodes omitted.

\end{document}